\documentclass[10pt,twocolumn,letterpaper]{article}

\usepackage{iccv}
\usepackage{times}
\usepackage{epsfig}
\usepackage{graphicx}
\usepackage{amsmath}
\usepackage{amssymb}
\usepackage{multirow}

\renewcommand{\a}{\mathbf{a}}
\newcommand{\Dt}{\Delta\boldsymbol{\tau}}
\newcommand{\x}{\mathbf{x}}

\newcommand{\g}{\mathbf{g}}
\newcommand{\h}{\mathbf{h}}
\newcommand{\y}{\mathbf{y}}

\renewcommand{\P}{\mathbf{P}}
\renewcommand{\ll}{\boldsymbol{\zeta}}

\newcommand{\qsection}[1]{\vspace{5pt} \noindent \textbf{#1:}}
\renewcommand{\Re}{\mathbb{R}}
\newcommand{\hf}{\hat{\mathbf{h}}}
\newcommand{\gf}{\hat{\mathbf{g}}}

\newcommand{\af}{\hat{\mathbf{a}}}
\newcommand{\xf}{\hat{\mathbf{x}}}
\newcommand{\yf}{\hat{\mathbf{y}}}
\newcommand{\Xf}{\hat{\mathbf{X}}}

\newcommand{\conj}{\mbox{conj}}
\newcommand{\lf}{\hat{\boldsymbol{\zeta}}}

\newcommand{\F}{\mathbf{F}}

\newcommand{\diag}{\mbox{diag}}

\usepackage[pagebackref=true,breaklinks=true,letterpaper=true,colorlinks,bookmarks=false]{hyperref}
\iccvfinalcopy 

\def\iccvPaperID{xxx} 

\ificcvfinal\fi

\begin{document}

\title{Learning Background-Aware Correlation Filters for Visual Tracking}

\author{%
\begin{tabular}[t]{c@{\extracolsep{4em}}c}
   \multicolumn{2}{c}{Hamed Kiani Galoogahi$^1$\thanks{Author Contact: \href{mailto:hamedkg@gmail.com}{hamedkg@gmail.com}}, Ashton Fagg$^{2,1}$, and Simon Lucey$^{1,2}$}\\
   \rule{0pt}{3ex}
   $^1$Robotics Institute & $^2$ SAIVT Lab\\ 
   Carnegie Mellon University, USA & Queensland University of Technology, Australia\\
\end{tabular}
}

\maketitle

\begin{abstract}
  Correlation Filters (CFs) have recently demonstrated excellent performance in terms of rapidly tracking objects under challenging photometric and geometric variations. The strength of the approach comes from its ability to efficiently learn - “on the fly” - how the object is changing over time. A fundamental drawback to CFs, however, is that the background of the target is not be modeled over time which can result in suboptimal performance. Recent tracking algorithms have suggested to resolve this drawback by either learning CFs from more discriminative deep features (\eg DeepSRDCF~\cite{danelljan2015convolutional} and CCOT~\cite{danelljan2016beyond}) or learning complex deep trackers (\eg MDNet~\cite{nam2015learning} and FCNT~\cite{wang2015visual}). While such methods have been shown to work well, their use comes at a high cost: extracting deep features or applying deep tracking frameworks is very computationally expensive. This limits the real-time performance of such methods, even on high-end GPUs. This work proposes a Background-Aware CF based on hand-crafted features (HOG~\cite{hog}) that can efficiently model how both the foreground and background of the object varies over time. Our approach, like conventional CFs, is extremely computationally efficient- and extensive experiments over multiple tracking benchmarks demonstrate the superior accuracy and real-time performance of our method compared to the state-of-the-art trackers including those based on a deep learning paradigm. 
\end{abstract}

\section{Introduction}

Correlation Filters (CFs) have been a widely used framework for visual object tracking~\cite{liang2015encoding,wu2015object,danelljan2015learning,henriques2015high}, due to their superior computation and fair robustness to photometric and geometric variations. CF trackers can learn and detect quickly in the frequency domain~\cite{Kumar2005}, being the most notable example the MOSSE tracker with the tracking speed of $\sim$700 frames per second~\cite{bolme2010visual}. Furthermore, these trackers learn ``on-the-fly''. The approach quickly models how an object varies visually over time by updating the tracker when the next frames become available. Such per frame adaptation offers robust tracking under challenging circumstances such as motion blur, scaling and lighting variation. 

Learning CF trackers in the frequency domain, however, comes at the high cost of learning from circular shifted examples of the foreground target. These shifted patches are implicitly generated through the circulant property of correlation in the frequency domain and are used as negative examples for training the filter~\cite{Kumar2005}. All shifted patches are plagued by circular boundary effects and are not truly representative of negative patches in real-world scenes~\cite{kiani2015correlation}.

These boundary effects have been shown to have a drastic impact on tracking performance, due to a number of factors. First, learning from limited shifted patches may lead to training an over-fitted filter which is not well-generalized to rapid visual deformation \eg caused by fast motion~\cite{danelljan2015learning}. Second, the lack of real negative training examples can drastically degrade the robustness of such trackers against cluttered background, and as a result, increase the risk of tracking drift specifically when the target and background display similar visual cues. Third, discarding background information from the learning process may reduce the tracker's ability to distinguish the target from occlusion patches. This limits the potential of such trackers to re-detect after an occlusion or out-of-plane movement ~\cite{danelljan2015learning}.

\begin{figure}
    \begin{center}

    \begin{tabular}{@{}c@{}@{}c@{}}
       \includegraphics[scale=.35]{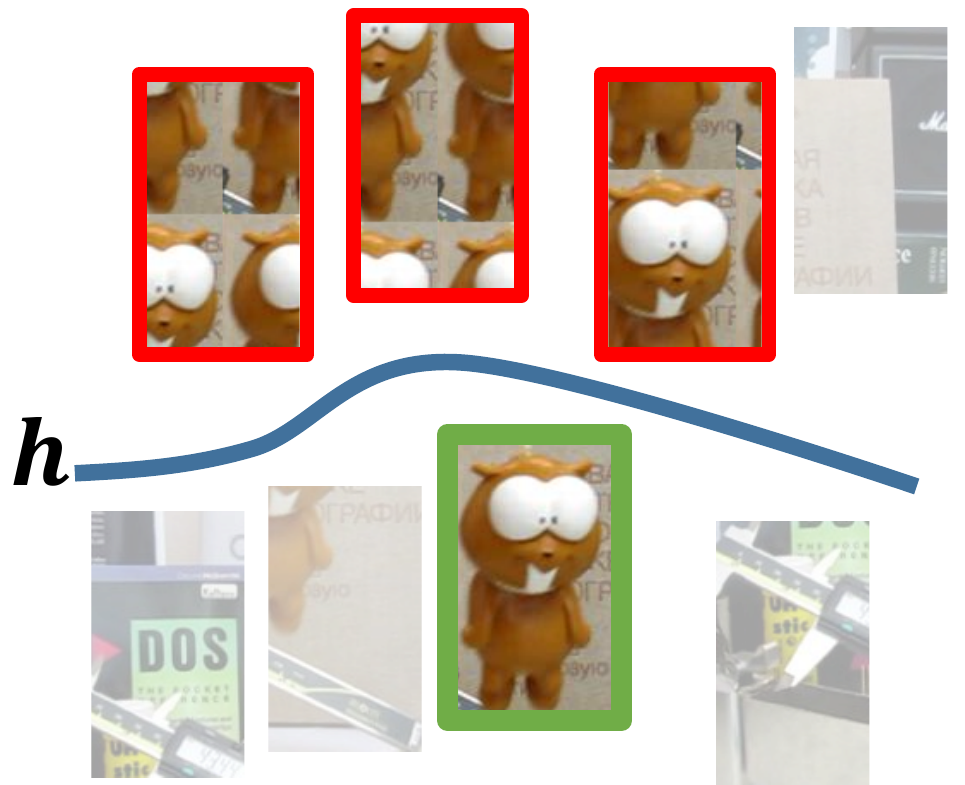}    &     \includegraphics[scale=.35]{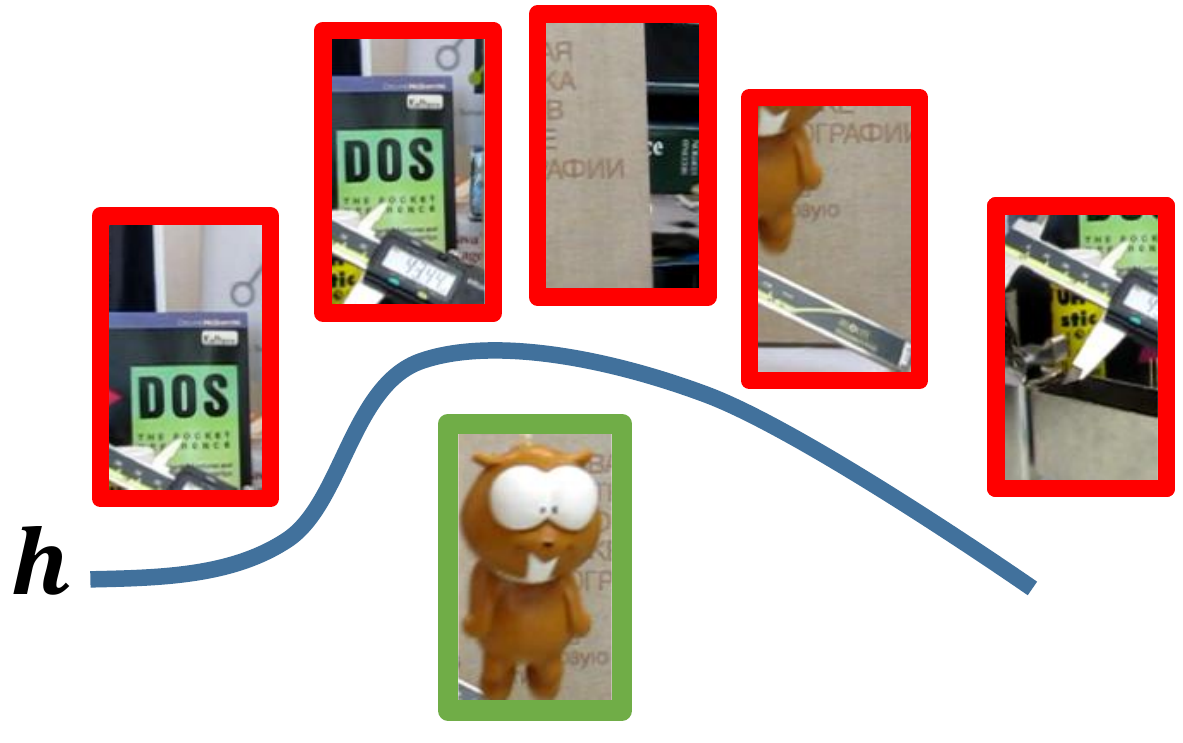}      \\
      (a)    &     (b)    
\end{tabular}
\end{center}
            \caption{(a) Traditional CFs discard the background patches, and instead learn from shifted patches of the cropped target. This may result to suboptimal results. (b) The BACF, however, exploits all background patches as negative examples for learning a filter which is more discriminative to background clutter. }
\label{fig:intro_fig}
\end{figure}

Recently, two methods were proposed to address the disadvantage of learning from shifted foreground patches~\cite{kiani2015correlation,danelljan2015learning}. The method of CFs with limited boundaries (CFLB) proposed to learn CFs with less boundary effects for the tasks of facial landmark localization and object tracking. Despite its promising results, this method was limited to learning CFs from pixel intensities- which as shown in~\cite{kiani2013multi} are not expressive enough for detecting challenging patterns in visual contents. Similar to our work, spatially regularized CFs (SRDCF)~\cite{danelljan2015learning} proposed to learn trackers from training examples with large spatial supports. The major disadvantage of this method is that the regularized objective is costly to optimize, even in the Fourier domain. Furthermore, in order to form the regularization weights, a set of hyper-parameters must be carefully tuned, which if not performed correctly can lead to poor tracking performance.


\qsection{Contribution} We propose to learn Background-Aware Correlation Filters (BACF) for real-time object tracking. Our method is capable of learning/updating filters from real negative examples densely extracted from the background. We demonstrate that learning trackers from negative background patches, instead of shifted foreground patches, achieves superior accuracy with real-time performance. This paper offers the following contributions:

\begin{itemize}
    \item We propose a new correlation filter for real-time visual tracking. Unlike prior CF-based trackers in which negative examples are limited to circular shifted patches, our tracker is trained from real negative training examples, densely extracted from the background.
   
    \item We propose an efficient Alternating Direction Method of Multipliers (ADMM) based approach for learning our filter on multi-channel features (\eg HOG), with computational cost of $\mathcal{O}(LKT \log(T))$, where $T$ is the size of vectorized frame, $K$ is the number of feature channels, and $L$ is the ADMM's iterations. 
    

\end{itemize}

We calculate model updates with Sherman-Morrison lemma to cope with changes in target and background appearance with real-time performance. We extensively evaluate our tracker on OTB50, OTB100, Temple-Color128 and VOT2015 datasets. The result demonstrates very competitive accuracy of our method compared to the state-of-the-art CF based and deep trackers, with superior real-time tracking speed of $\sim40$ FPS on a CPU.

 \begin{figure*}
    \begin{center}
         \begin{tabular}{c}
      
          \includegraphics[width=0.75\textwidth]{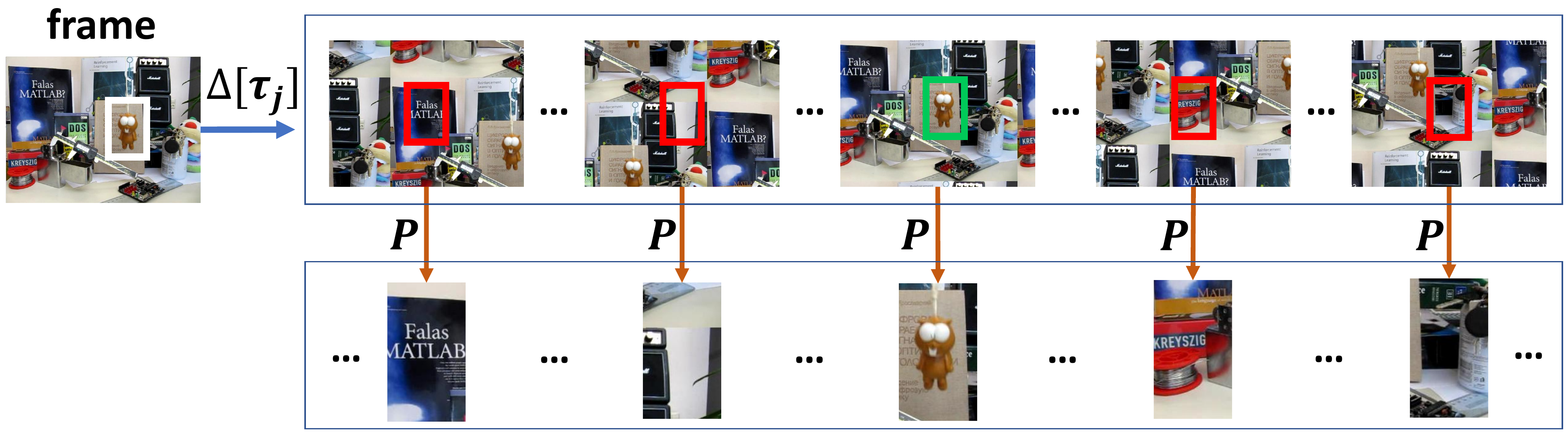}
 
             \end{tabular}
     \end{center}
            \caption{BACF learns from all possible positive and negative patches extracted from the entire frame. $[\Dt_{j}]$ generates all circular shifts of the frame over all $j= [0, ... , T-1]$ steps. $T$ is the length of vectorized frame. $\P$ is the cropping operator (a binary matrix) which crops the central patch of each shifted image. The size of the cropped patches is same as the size of the target/filter ($D$), where $T \gg D$. All cropped patches are utilized to train a CF tracker. In practice, we \textit{do not} apply circular shift and cropping operators. Instead, we perform these operations efficiently by augmenting our objective in the Fourier domain. The red and green boxes indicate the negative (background) and positive (target) training patches.  Please refer to Section \hyperref[sec:bacf]{4} for more details. }
\label{Fig:bacf_operation}
\end{figure*}

\section{Prior Art} The interest in employing CFs for visual tracking was ignited by the seminal work of Bolme~\etal~\cite{bolme2010visual} on the MOSSE filter with an impressive speed of $\sim$700 FPS. Thereafter, several works~\cite{danelljan2014accurate,danelljan2016adaptive,henriques2015high,li2014scale,bertinetto2015staple} were built upon the MOSSE approach showing notable improvement by learning CFs trackers on multi-channel features such as HOG~\cite{hog}. All these approaches, however, imitated the standard formulation of CFs in the frequency domain to retain their computations efficient for real-time applications. Learning CF trackers quickly in the frequency domain, however, is highly affected by boundary effects of shifted patches~\cite{kiani2015correlation}, leading to suboptimal training~\cite{danelljan2015learning}. Moreover, such methods solely learn trackers from object patches cropped from the whole frame, and the background visual information is discarded from the learning process. This leads to poor discrimination against cluttered background, and thereby, increases the risk of spurious detection when the target and its surrounding background share similar visual cues~\cite{danelljan2015learning}. Several recent works addressed the constraint of learning from shifted patches by exploiting training samples whose spatial size is much larger than the trained filters~\cite{kiani2015correlation,danelljan2015learning,danelljan2016adaptive,danelljan2015convolutional}. Learning from large training samples not only dramatically reduces boundary effects~\cite{kiani2015correlation}, but offers learning filters from a huge number of background patches~\cite{danelljan2015learning}. The method of CFLB~\cite{kiani2015correlation} was originally designed to learn from pixel intensities which was shown to be inaccurate and suffers from poor generalization on challenging patterns~\cite{kiani2013multi}. Our method, on the other hand, is capable of handling more discriminative and well-generalized multi-channel features such as HOG~\cite{hog} and convolutional neural networks (CNNs) features~\cite{simonyan2014very}. The SRDCF method~\cite{danelljan2015learning} and its variations~\cite{danelljan2016adaptive,danelljan2015convolutional} require regularization weights to penalize the correlation filter coefficients during learning. These weights are highly target and video dependent, and have to be carefully fine-tuned over a set of sensitive hyper-parameters to perform well for each video. Furthermore, due to their computational expense ($\sim$4 FPS), SRDCF methods are not a suitable choice for real-time tracking. 

The excellent performance of deep convolutional neural networks (CNNs) on several challenging vision tasks~\cite{krizhevsky2012imagenet,simonyan2014very,long2015fully,jia2014caffe} has encouraged more recent works to either exploit CNN deep features within CFs framework~\cite{ma2015hierarchical, danelljan2015convolutional, danelljan2016beyond} or design deep architectures~\cite{bertinetto2016fully,nam2015learning,wang2016stct,chen2016once,tao2016siamese,wang2015visual} for robust visual tracking. This trend has its own pros and cons. Compared to hand-crafted features such as HOG, learning CF trackers using CNN features significantly improves their robustness against geometric and photometric variations~\cite{danelljan2016beyond}. This is mainly resulted from the high discrimination of such features, since CNNs are trained over large scale dataset~\cite{krizhevsky2012imagenet}. However, extracting CNN features from each frame and training/updating CF trackers over high dimensional deep features is computationally expensive. Such an approach leads to poor real time performance ($\sim$~0.2 FPS in the case of ~\cite{danelljan2015convolutional, danelljan2016beyond}). Similarly, purely deep trackers also suffer the same drawback ~\cite{bertinetto2016fully,nam2015learning,wang2016stct,chen2016once, tao2016siamese,wang2015visual}, with some methods performing at only 1 FPS on a typical desktop PC.


\section{Correlation Filters}

Learning multi-channel CFs in the spatial domain is formulated by minimizing the following objective~\cite{kiani2013multi}:

\begin{equation}
E(\h)  =  \frac{1}{2} || \y -
\sum_{k=1}^{K} \h_{k} \star \x_{k} ||_{2}^{2}
+ \frac{\lambda}{2} \sum_{k=1}^{K}||\h_{k}||_{2}^{2} \label{Eq:MultiChannel_corr}
\end{equation}

where~$\x_{k} \in \Re^{D}$ and~$\h_{k} \in \Re^{D}$ refers to the~$k$th channel of the vectorized image and filter respectively, and~$K$ is the number of feature channels.  ~$\y \in \Re^{D}$ is the desired correlation response,~$\lambda$ is a regularization, and $\star$ is the spatial correlation operator. Eq.~\ref{Eq:MultiChannel_corr} can be identically expressed as a ridge regression objective in the spatial domain:

\begin{eqnarray}
E(\h) = & &\frac{1}{2} \sum_{j =1}^{D} || \y(j) -
\sum_{k=1}^{K} \h^\top_{k} \x_{k}[\Dt_{j}] ||_{2}^{2}  \nonumber \\
 & &+ \quad \frac{\lambda}{2} \sum_{k=1}^{K}||\h_{k}||_{2}^{2} \label{Eq:MultiChannel}
\end{eqnarray}

where $\y(j)$ is the $j$-th element of $\y$. $[\Dt_{j}]$ is the circular shift operator, and $\x_{k}[\Dt_{j}]$ applies a $j$-step discrete circular shift to the signal $\x_{k}$. For a full treatment of multi-channel correlation filters, please see~\cite{kiani2013multi}.

The main drawback of Eq.~\ref{Eq:MultiChannel} is learning a correlation filter/detector from $D-1$ circular shifted foreground patches which are generated through the $[\Dt_{j}]$ operator. This trains a filter which perfectly discriminates the foreground target from its shifted examples. As mentioned earlier, this, however, increases the risk of over-fitting and limits the potential of the filter to classify the target from real non-target patches (Fig.~\ref{fig:intro_fig} (a)). For generic object detection task (such as pedestrian detection in~\cite{kiani2013multi}), this drawback can be significantly diminished by exploiting a huge amount of positive (pedestrian) and negative (non-pedestrian) patches to train a well-generalized filter/detector. This, however, is not practical for the task of visual tracking. The target is the only positive sample available at the training time and gathering positive and negative examples from a pre-collected training set for each individual target is infeasible. Fortunately, the target comes with a large surrounding background which can be used as negative samples at the training stage. We propose the method of background-aware correlation filters to directly learn more robust and well-generalized CF tracker from background patches (Fig.~\ref{fig:intro_fig} (b)).

\section{Background-Aware Correlation Filters}
\label{sec:bacf}
We propose to learn multi-channel background-aware correlation filters by minimizing the following objective:

\begin{eqnarray}
E(\h) & = & \frac{1}{2} \sum_{j =1}^{T} || \y(j) -
\sum_{k=1}^{K} \h^\top_{k} \textbf{P} \x_{k}[\Dt_{j}] ||_{2}^{2} \nonumber \\
 & & + \quad \frac{\lambda}{2} \sum_{k=1}^{K}||\h_{k}||_{2}^{2} \label{Eq:MCCFLB}
\end{eqnarray}

where $\textbf{P}$ is a $D \times T$ binary matrix which crops the mid $D$ elements of signal $\x_{k}$. In this formulation, $\x_{k} \in \Re^{T}$, $\y \in \Re^{T}$ and $\h \in \Re^{D}$, where $T \gg D$.

For tracking task, $\x$, $\y$, and $\h$ are respectively a training sample with large spatial support, $\y$ is the correlation output with a peak centered upon the target of interest, and $\h$ is the correlation filter whose spatial size is much smaller than training samples. Applying the circular shift operator on the training sample followed by the cropping operator, $\textbf{P}\x_{k}[\Dt_{j}]$, returns all possible patches with the size of $D$ from the entire frame, Fig.~\ref{Fig:bacf_operation}. The cropped patch corresponding to the peak of the correlation output displays the target (positive example), and those corresponding to the zero values of the correlation output display the background content (negative examples). The computational cost of Eq.~\ref{Eq:MCCFLB} is approximately the same as Eq.~\ref{Eq:MultiChannel}, $\mathcal{O}(D^{3}K^{3})$, since $\textbf{P}$ can be precomputed, and $\textbf{P} \x_{k}$ (cropping) can be efficiently performed via a lookup table.  

Correlation filters are typically learned in the frequency domain, for computational efficiency ~\cite{Kumar2005}. Similarly, Eq.~\ref{Eq:MCCFLB} can be expressed in the frequency domain as:

\begin{eqnarray}
E(\h,\gf) & = & \frac{1}{2} || \yf - \Xf \gf ||_{2}^{2}
+ \frac{\lambda}{2} ||\h||_{2}^{2} \nonumber \\
& \mbox{s.t.} & \gf = \sqrt{T}(\F\P^{\top} \otimes \mathbf{I}_K)\h
\label{Eq:mccflb_e}
\end{eqnarray}

where, $\gf$ is an auxiliary variable and the matrix $\Xf$ is defined as~$\Xf = [\diag(\xf_{1})^{\top},\ldots,\diag(\xf_{K})^{\top}]$ of size $T \times KT$. $\h = [\h_{1}^{\top},\ldots,\h_{K}^{\top}]^{\top}$ and $\gf = [\gf_{1}^{\top},\ldots,\gf_{K}^{\top}]^{\top}$ respectively show the $KD \times 1$ and $KT \times 1$ over-complete representations of $\h$ and $\gf$ by concatenating their $K$ vectorized channels. $\mathbf{I}_K$ is a $ K \times K$ identity matrix, and $\otimes$ indicates the Kronecker product. A~$\hat{}$~denotes the Discrete Fourier Transform (DFT) of a signal, such that~$\af  = \sqrt{T} \F \a$, where $\F$ is the orthonormal~$T \times T$ matrix of complex basis vectors for mapping to the Fourier domain for any~$T$ dimensional vectorized signal. The transpose operator~$^{\top}$ on a complex vector or matrix computes the conjugate transpose.

\subsection{Augmented Lagrangian}

To solve Eq.~\ref{Eq:mccflb_e}, we employ an Augmented Lagrangian Method (ALM)~\cite{boyd_book_2010}:

\begin{eqnarray} \label{mccflc_al}
\mathcal{L}(\gf,\h,\lf) & = & \frac{1}{2} || \yf -
 \Xf  \gf ||_{2}^{2} + \frac{\lambda}{2} ||\h||_{2}^{2} \nonumber \\
& & + \quad \lf^{\top}(\gf - \sqrt{T} (\F \P^{\top} \otimes \mathbf{I}_K) \h ) \nonumber \\
& & + \quad \frac{\mu}{2} ||\gf - \sqrt{T} (\F \P^{\top} \otimes \mathbf{I}_K) \h ||_{2}^{2}
\end{eqnarray}

where~$\mu$ is the penalty factor and~$\lf = [\lf_{1}^{\top},\ldots,\lf_{K}^{\top}]^{\top}$ is the $KT \times 1$ Lagrangian vector in the Fourier domain. Equation~\ref{mccflc_al} can be solved iteratively using the ADMM \cite{boyd_book_2010} technique. Each of the subproblems, $\hat{\mathbf{g}}^*$ and $\mathbf{h}^*$, have closed form solutions.

\qsection{Subproblem $\mathbf{h}^*$}

\begin{eqnarray}\label{Eq:mccf_lb_h} \nonumber
\h^{*} & = & \arg \min_{\h} \Bigl\{ \frac{\lambda}{2} ||\h||_{2}^{2} +  \lf^{\top}(\gf - \sqrt{T} (\F \P^{\top} \otimes \mathbf{I}_K) \h ) \\ \nonumber &&  + \quad \frac{\mu}{2} ||\gf - \sqrt{T} (\F \P^{\top} \otimes \mathbf{I}_K) \h ||_{2}^{2} \Bigr\} \\
          & = & (\mu + \frac{\lambda}{\sqrt{T}})^{-1}(\mu \g + \ll)  
\end{eqnarray}

where~$\g = \frac{1}{\sqrt{T}}(\P\F^{\top}\otimes \mathbf{I}_K)\gf$ and ~$\ll = \frac{1}{\sqrt{T}}(\P\F^{\top}\otimes \mathbf{I}_K)\lf$. The Kronecker product with the identity matrix can be broken into $K$ independent Inverse Fast Fourier Transform (IFFT) computations of $\g_{k} = \frac{1}{\sqrt{T}} \P\F^{\top} \gf_{k}$ and $\ll_{k} = \frac{1}{\sqrt{T}} \P\F^{\top} \lf_{k}$. In practice, both~$\g_{k}$ and $\ll_{k}$ can be estimated efficiently by applying an IFFT on each~$\gf_{k}$ and $\lf_{k}$ and then applying the lookup table formed from the masking matrix $\P$. The over-complete vectors $\g$ and $\ll$ can be obtained by concatenating ~$\{ \g_{k} \}_{k=1}^{K}$ and $\{ \ll_{k} \}_{k=1}^{K}$, respectively. The computation of Eq.~\ref{Eq:mccf_lb_h} is bounded by $\mathcal{O}(K  T \log (T))$, where $K$ is the number of channels, and $T \log (T)$ is the cost of computing the IFFT of a signal with the length of $T$.

\qsection{Subproblem $\hat{\mathbf{g}}^*$}

\begin{eqnarray} \label{Eq:mccf_lb_g}  
\gf^{*} & = & \arg \min_{\gf} \Bigl\{    \frac{1}{2} || \yf -
 \Xf  \gf ||_{2}^{2}  \nonumber \\
& & + \quad \lf^{\top}(\gf - \sqrt{T} (\F \P^{\top} \otimes \mathbf{I}_K) \h ) \nonumber \\
& & + \quad \frac{\mu}{2} ||\gf - \sqrt{T} (\F \P^{\top} \otimes \mathbf{I}_K) \h ||_{2}^{2}  \Bigr\}  
\end{eqnarray}

Solving Eq.~\ref{Eq:mccf_lb_g} directly is $\mathcal{O}(T^{3}K^{3})$. This computation is intractable for real-time tracking, since we need to solve for $\hat{\mathbf{g}}^*$ at every ADMM iteration. Fortunately, $\Xf$ is sparse banded, and thus, each element of $\yf$ ($\yf(t), ~t = 1,...,T$) is dependent only on $K$ values of $\xf(t) = [\xf_{1}(t), ... , \xf_{K}(t)]^{\top}$ and $\gf(t) = [\conj{(\gf_{1}(t))}, ... ,\conj{(\gf_{K}(t))}]^{\top}$~\cite{kiani2013multi}. The operator $\conj{(.)}$ applies the complex conjugate to a complex vector/number. Therefore, solving Eq.~\ref{Eq:mccf_lb_g} for $\gf^{*} $ can be identically expressed as $T$ smaller, independent objectives, solving for $\gf(t)^{*}$,  over  $t = [1, ..., T]$:

\begin{eqnarray} \label{Eq:mccf_lb_g_smaller}  
\gf(t)^{*} & = & \arg \min_{\gf(t)} \Bigl\{    \frac{1}{2} || \yf (t)-
 \xf(t)^\top  \gf(t) ||_{2}^{2}  \nonumber \\
& & + \quad \lf(t)^{\top}(\gf(t) -  \hf(t) ) \nonumber \\
& & + \quad \frac{\mu}{2} ||\gf(t) - \hf(t) ||_{2}^{2}  \Bigr\} 
\end{eqnarray}

where $\hf(t) = [\hf_{1}(t), ... , \hf_{K}(t)]$ and $\hf_{k} = \sqrt{D}\F\P^{\top} \h_{k}$. In practice, each ~$\hf_{k}$ can be estimated efficiently by applying a FFT to each ~$\h_{k}$ padded with zeros. The solution for each $\gf(t)^{*} $ is obtained by:


\begin{multline}
\gf(t)^{*} = \left( \xf(t) \xf(t)^\top + T\mu \mathbf{I}_K \right)^{-1}\\\left(\yf(t)\xf(t) - T\lf(t) + T\mu \hf(t) \right)
\label{Eq:mccf_lb_g_smaller_sol}
\end{multline}

Eq.~\ref{Eq:mccf_lb_g_smaller_sol} has the complexity of $\mathcal{O}(TK^{3})$, since we still need to solve $T$ independent $K \times K$ linear systems. Even though this computation is substantially smaller than directly solving ($\mathcal{O}(T^{3}K^{3})$), it is still intractable for real-time tracking.

\qsection{Real-Time Extension} We propose to compute $\left( \xf(t) \xf(t)^\top + T \mu \mathbf{I}_K \right)^{-1}$ rapidly using the Sherman-Morrison formula~\cite{sherman1950adjustment}, stating that $(\mathbf{u}\mathbf{v}^{\top} + \mathbf{A})^{-1} =  \mathbf{A}^{-1} - (\mathbf{v}^{\top} \mathbf{A}^{-1}\mathbf{u})^{-1} \mathbf{A}^{-1}\mathbf{u}\mathbf{v}^{\top} \mathbf{A}^{-1}$, where in our case, $ \mathbf{A} = T\mu\mathbf{I}_K$ and $\mathbf{u} = \mathbf{v} = \xf(t)$. Hence, Eq.~\ref{Eq:mccf_lb_g_smaller_sol} can be rewritten as:

\begin{eqnarray} \label{Eq:mccf_lb_g_smaller_sol_SM}  
\gf(t)^{*} &=& \frac{1}{\mu}\left(T\yf(t)\xf(t) -\lf(t)+\mu\hf(t) \right) \\
&&-\frac{\xf(t)}{\mu b}\left(T\yf(t)\hat{s}_{\x}(t) - \hat{s}_{\ll}(t) + \mu\hat{s}_{\h}(t)\right)  \nonumber
\end{eqnarray}
where, $\hat{s}_{\x} (t) = \xf(t)^{\top}\xf $, $\hat{s}_{\ll} (t) = \xf(t)^{\top}\lf $, $\hat{s}_{\h} (t) = \xf(t)^{\top}\hf $ and $b =\hat{s}_{\x} (t) + T\mu$ are scalar. The cost of computing $\gf$ using Eq.~\ref{Eq:mccf_lb_g_smaller_sol_SM} is $\mathcal{O}(TK)$, whichi is much smaller than the computation of Eq.~\ref{Eq:mccf_lb_g_smaller_sol} ($\mathcal{O}(TK^{3})$).

\qsection{Lagrangian Update}
We update the Lagrangians as
\begin{equation}
\lf^{(i+1)} \leftarrow \lf^{(i)} + \mu(\gf^{(i+1)} - \hf^{(i+1)})
\label{Eq:mccf_lb_l}
\end{equation}
where~$\gf^{(i+1)}$ and~$\hf^{(i+1)}$ are the current solutions to the above subproblems at iteration~$i+1$ within the iterative ADMM, and $\hf^{(i+1)} = (\F \P^{\top} \otimes \mathbf{I}_K) \h^{(i+1)}$. A common scheme for selecting~$\mu$ is $\mu^{(i+1)} = \min(\mu_{\max},\beta \mu^{(i)})$~\cite{boyd_book_2010}.

\qsection{Online Update} Similar to other CF trackers~\cite{henriques2015high,bolme2010visual,bertinetto2015staple}, we utilize an online adaptation strategy to improve our robustness to pose, scale and illumination changes. The online adaptation at frame $f$ is formulated as $\xf^{(f)}_{model} = (1 - \eta) \quad \xf^{(f-1)}_{model} + \eta \quad \xf^{(f)}$, where $\eta$ is the online adaptation rate. Based on this strategy, we use $\xf^{(f)}_{model}$ instead of $\xf^{(f)}$ in Eq.~\ref{Eq:mccf_lb_g_smaller_sol_SM} to compute $\gf(t)^{*}$, $\hat{s}_{\x} (t)$, $\hat{s}_{\ll} (t)$ and $\hat{s}_{\h}$.

\qsection{Detection} The spatial location of the target in frame $f$ is detected by applying the filter $\gf^{(f-1)}$ that has been updated in the previous frame. Following~\cite{danelljan2015learning,bertinetto2015staple}, the filter is applied on multiple resolutions of the searching area to estimate scale changes. The searching area has the same spatial size of the filter $\gf$. This returns $S$ correlation outputs, where $S$ is the number of scales. We employ the interpolation strategy in~\cite{danelljan2015learning,danelljan2016beyond} to maximize detection scores per each correlation output. The scale with the maximum correlation output is used to update the object location and scale.  

\section{Experiments}
We extensively evaluate our tracker on four standard datasets, including OTB50~\cite{wu2013online}, OTB100~\cite{wu2015object}, Temple-Color128 (TC128)~\cite{liang2015encoding}, and VOT2015~\cite{kristan2015visual}, comparing with 24 state-of-the-art methods, such as TLD~\cite{kalal2012tracking},
Struck~\cite{hare2011struck}, CFLB~\cite{kiani2015correlation}, KCF~\cite{henriques2015high}, DSST~\cite{danelljan2014accurate}, SAMF~\cite{li2014scale}, MEEM~\cite{zhang2014meem}, DAT~\cite{possegger2015defense}, LCT~\cite{ma2015long}, HCF~\cite{ma2015hierarchical}, Staple~\cite{bertinetto2015staple}, SRDCF~\cite{danelljan2015learning}, SRDCFdecon~\cite{danelljan2016adaptive}, DeepSRDCF~\cite{danelljan2015convolutional}, CCOT~\cite{danelljan2016beyond}, S3Tracker~\cite{lee2011visual}, SC-EBT~\cite{wang2014ensemble}, LDP~\cite{lukevzivc2016deformable}, SiamFC~\cite{bertinetto2016fully}, MDNet~\cite{nam2015learning}, STCT~\cite{wang2016stct}, YCNN~\cite{chen2016once}, SINT~\cite{tao2016siamese} and FCNT~\cite{wang2015visual}.

\qsection{Evaluation Methodology} We use the success metric~\cite{wu2013online} to evaluate all trackers on OTB50, OTB100 and TC128. Success measures the intersection over union (IoU) of predicted and ground truth bounding boxes. The success plot shows the percentage of bounding boxes whose IoU score is larger than a given threshold. We use the Area Under the Curve (AUC) of success plots to rank the trackers. We also compare all the trackers by the success rate at the conventional thresholds of 0.50 (IoU $>$ 0.50)~\cite{wu2013online}. For the VOT15 dataset, tracking performance is evaluated in terms of accuracy (overlap with the ground-truth) and robustness (failure rate)~\cite{kristan2015visual}. In VOT2015, a tracker is restarted in the case of a failure, where there is no overlap between the detected bounding box and ground truth. For a full treatment of these metrics, readers are encouraged to read ~\cite{kristan2015visual,wu2013online}. 

\qsection{Comparison Scenarios} We evaluate the BACF tracker over four experiments. The first experiment is conducted to show the superiority of our method to the state-of-the-art trackers with hand-crafted features (HOG). At the second experiment, we compare BACF with CF trackers with deep features to demonstrate compared to such methods BACF offers very competitive accuracy with almost two orders of magnitude (170 times) improvement in tracking speed. The third experiment compares our method with the state of the art deep trackers, and the last experiment compares our tracker with leading methods of the VOT2015 challenge. 

\qsection{Implementation Details} Similar to recent CF trackers~\cite{danelljan2015learning,henriques2015high,bertinetto2015staple} we employ 31-channel HOG features~\cite{hog} using $4 \times 4$ cell size multiplied by a Hann window~\cite{bolme2010visual}. The regularization factor, $\lambda$, is set to 0.001, and the number of scales ($S$) is set to 5 with an scale-step of 1.01. A 2D Gaussian function with bandwidth of $\sqrt{wh}/16$ is used to define the correlation output for an object of size $[h,w]$. For the ADMM optimization, we set the number of iterations and the penalty factor, $\mu$, to 2 and 1, respectively. The penalty factor at iteration $i+1$ is updated by $\mu^{(i+1)} = \min(\mu_{\max},\beta \mu^{(i)})$, where $\beta = 10$ and $\mu_{\max} = 10^3$. We tested different configurations of $\mu$ and $\beta$, and we observed that choosing large values of $\beta$ and $\mu$ over very few iterations helps the ADMM to converge much faster than smaller $\beta$ and $\mu$ over more ADMM iterations. This substantially saves learning complexity with almost the same tracking accuracy. The learning (adaptation) rate of BACF $\eta = 0.0125$ for all experiments. We tested our MATLAB implementation on a machine equipped with an Intel Core i7  running at 2.40 GHz.

\begin{figure*}
    \begin{center}
         \begin{tabular}{@{}c@{} @{}c@{} @{}c@{} }
      {\footnotesize {\textbf{Success plot on OTB50}}} &
      {\footnotesize {\textbf{Success plot on OTB100}}} &
      {\footnotesize {\textbf{Success plot on TC128}}}  \\
          \includegraphics[scale=.465]{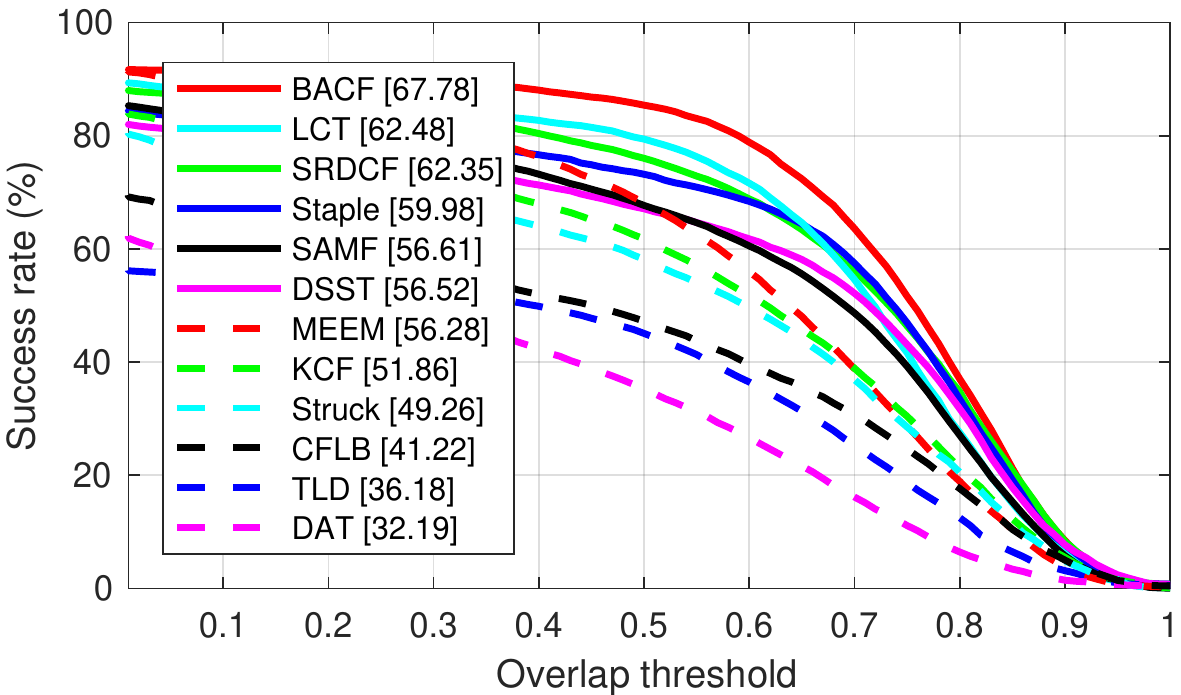} 
&  \includegraphics[scale=.465]{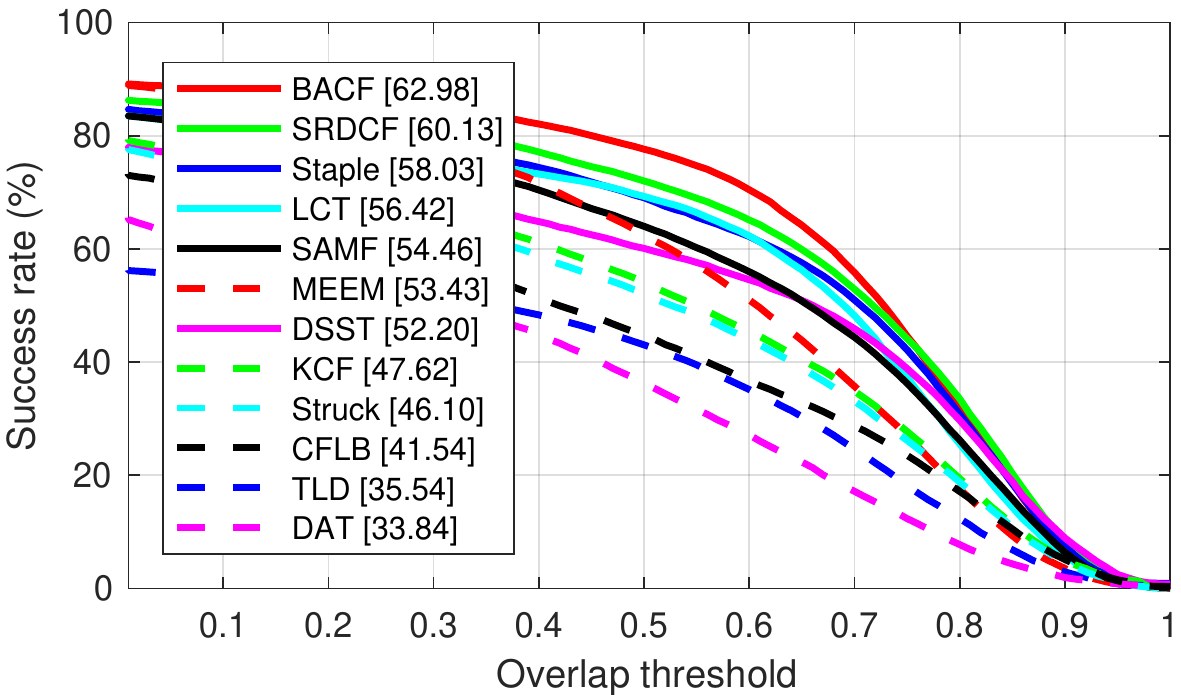} 
& \includegraphics[scale=.465]{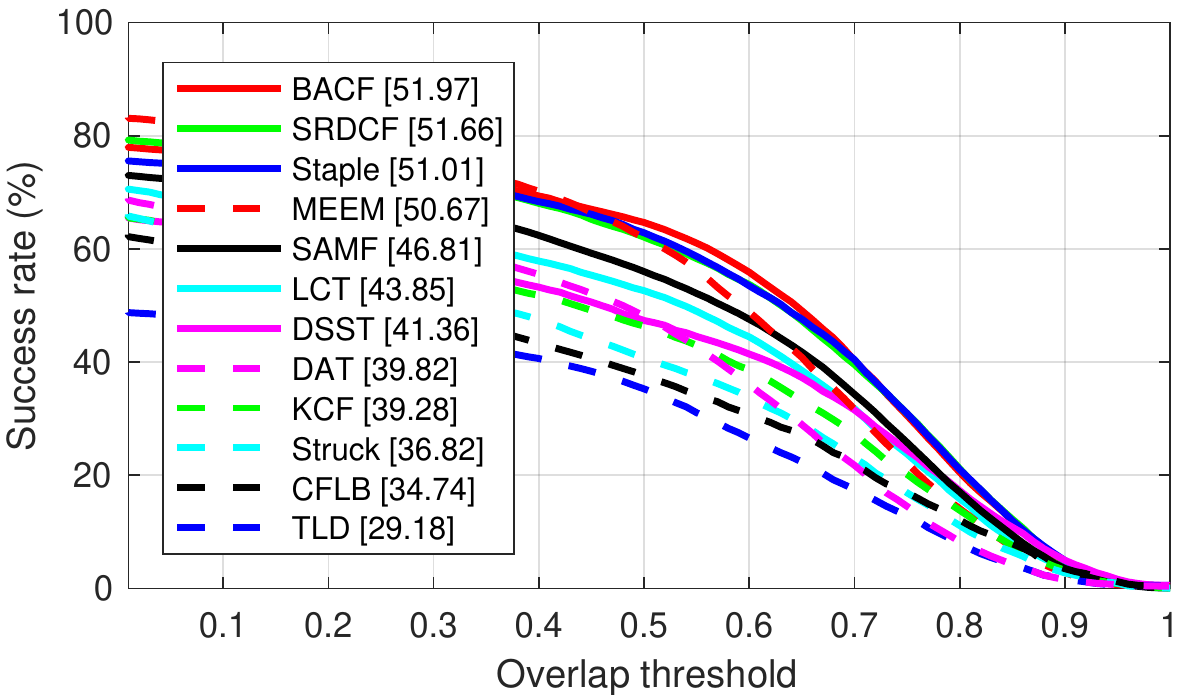} 

  \\
 (a) & (b) & (c) 
             \end{tabular}
     \end{center}
            \caption{Success plots comparing BACF with the state-of-the-art HOG based trackers on (a) OTB50, (b) OTB100, and (c) TC128. We also include CFLB as the latest pixel intensities-based CF tracker. AUCs are reported in brackets.}
\label{Fig:OP_evaluation}
\end{figure*}

\begin{table*}
\centering
\caption{Success rates ($\%$ at IoU $>$ 0.50) of BACF versus HOG-based trackers. The {\color{red}first}, {\color{green}second} and {\color{blue}third} best methods are shown in color.}
\label{Table:CF_HOG}
\begin{tabular}{lcccccccccccc}
\hline
                       & BACF & SRDCF & Staple & LCT & SAMF & MEEM & DSST & KCF & Struck & CFLB & TLD & DAT \\ \hline
\multicolumn{1}{l}{OTB50} &   {\color{red}85.4}   &    {\color{blue}76.0}   &    73.2    &  {\color{green}79.4}   & 67.7     &  68.2    &  67.1    &    61.8 &      58.2  &     47.3 &  45.1   &    35.2 \\  
\multicolumn{1}{l}{OTB100} &   {\color{red}77.6}   &   {\color{green}72.0}    &   69.1     &  {\color{blue}69.3}   &  64.0    &   62.6   &    60.1  &   54.2  &    52.2    &   44.7   &   43.1  &   36.3  \\  
\multicolumn{1}{l}{TC128} &   {\color{red}65.2}  &    {\color{blue}62.1}   &    {\color{green}62.9 }   &  52.6   &   56.0   &   62.0   &   47.4   &  46.4   &  40.7      &   37.7   &  35.3   &  48.1   \\ \hline \hline
\multicolumn{1}{l}{Avg. succ. rate} &   {\color{red}76.0 }  &   {\color{green}70.0 }   &     {\color{blue}68.4}   &   67.1  &   62.5   &  64.2    &   58.2   &   54.1  &     50.3   &  43.2    &  41.1   &  39.8   \\  
\multicolumn{1}{l}{Avg. FPS} &   35.3   &  3.8     &    48.3    &  18.5   &   11.4   &    11.1  &    17.7  &  {\color{red}173.4}   &    9.2    &   {\color{green}87.1}   &   22.1  &   {\color{blue}60.3}  \\ \hline
\end{tabular}
\end{table*}

\begin{figure*}
    \begin{center}
         \begin{tabular}{@{}c@{} @{}c@{} @{}c@{} }

       {\footnotesize {\textbf{Success plot of background clutter (31)}}} &
        {\footnotesize {\textbf{Success plot of scale variation (64)}}}&
         {\footnotesize {\textbf{Success plot of occlusion (49)}}} \\
      
          \includegraphics[scale=.45]{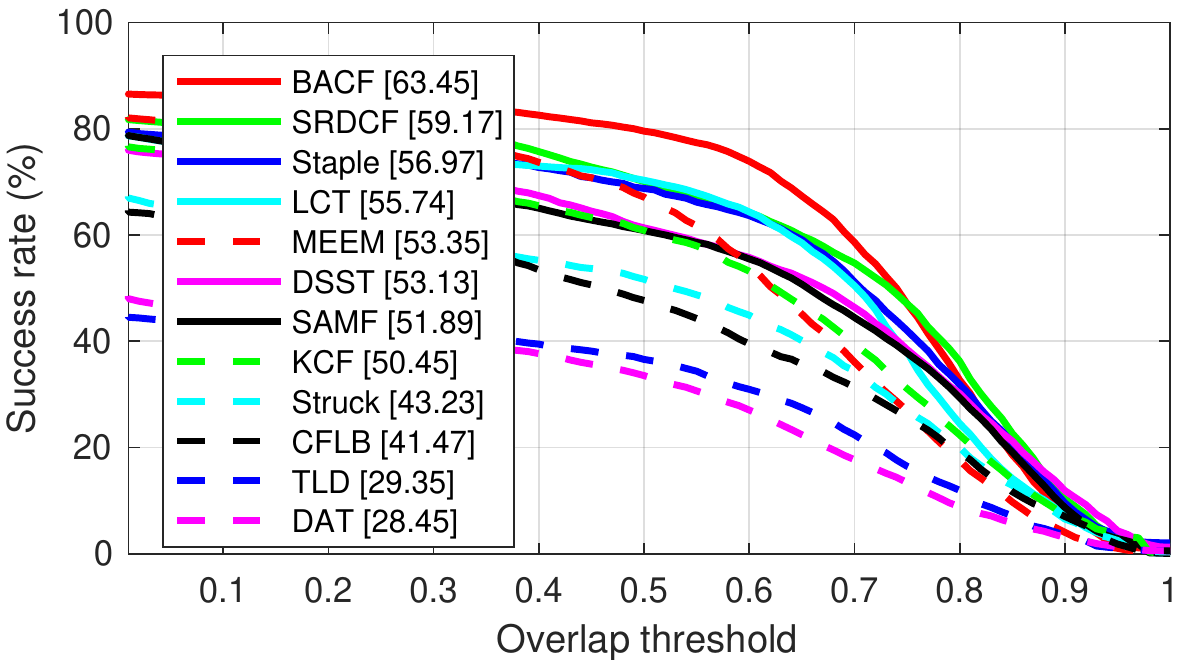} 
&  \includegraphics[scale=.45]{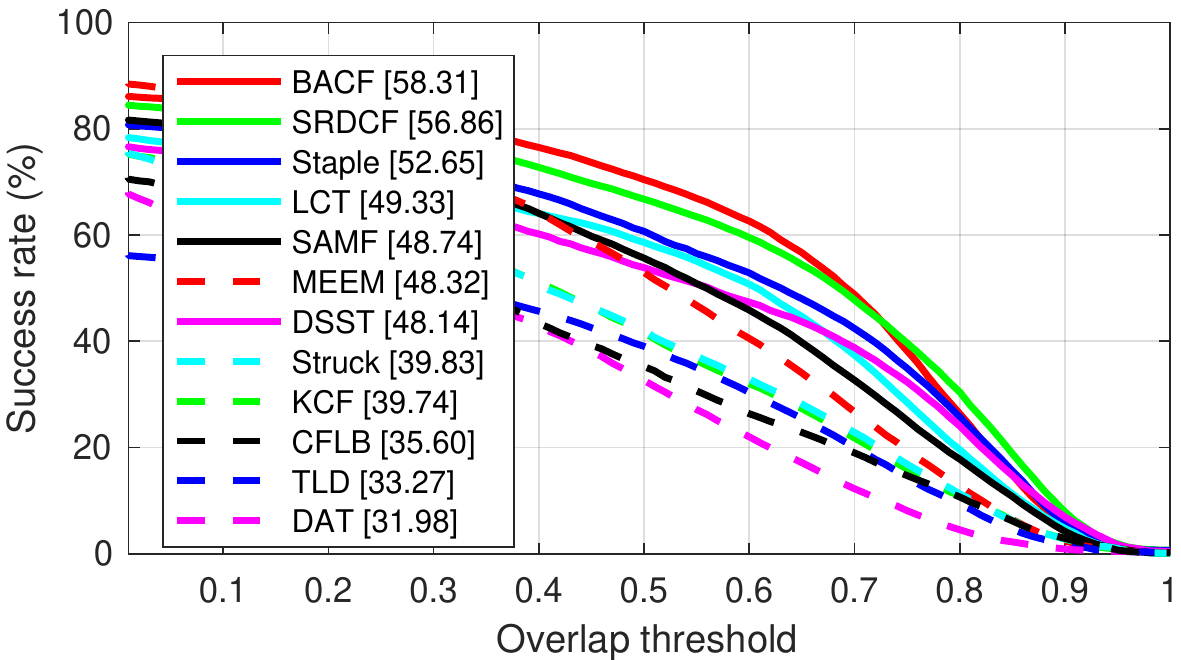} 
& \includegraphics[scale=.45]{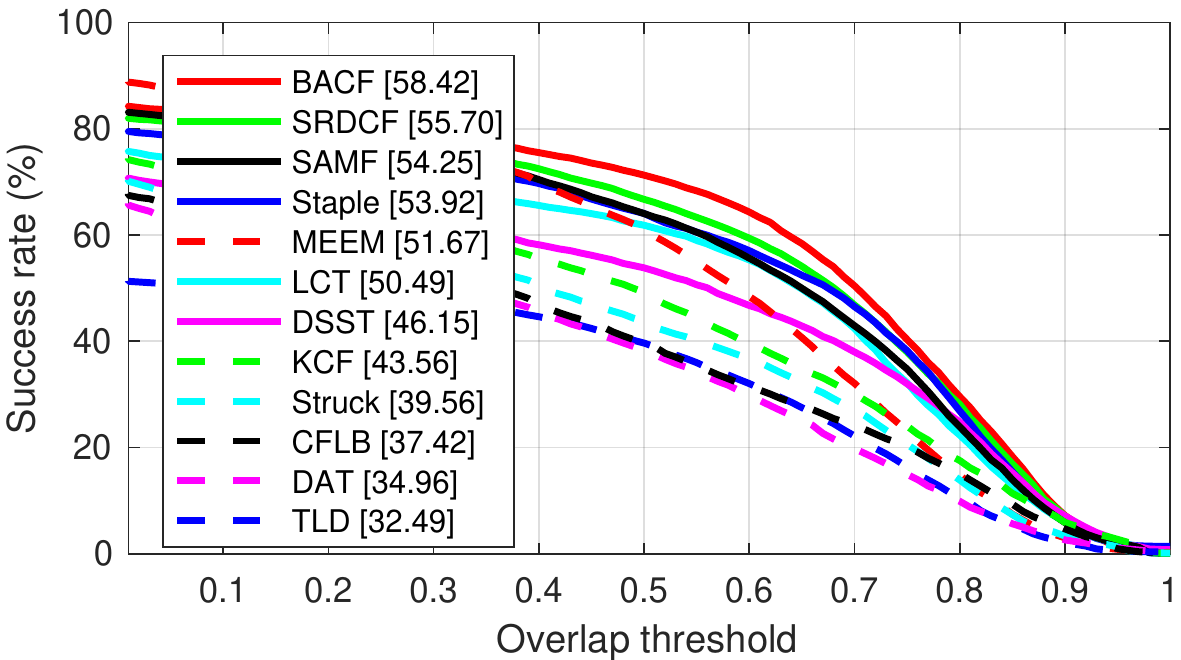} 
  \\
  
   {\footnotesize {\textbf{Success plot of deformation (44)}}} &
        {\footnotesize {\textbf{Success plot of motion blur (29)}}}&
         {\footnotesize {\textbf{Success plot of out of view (14)}}} \\
         
    \includegraphics[scale=.45]{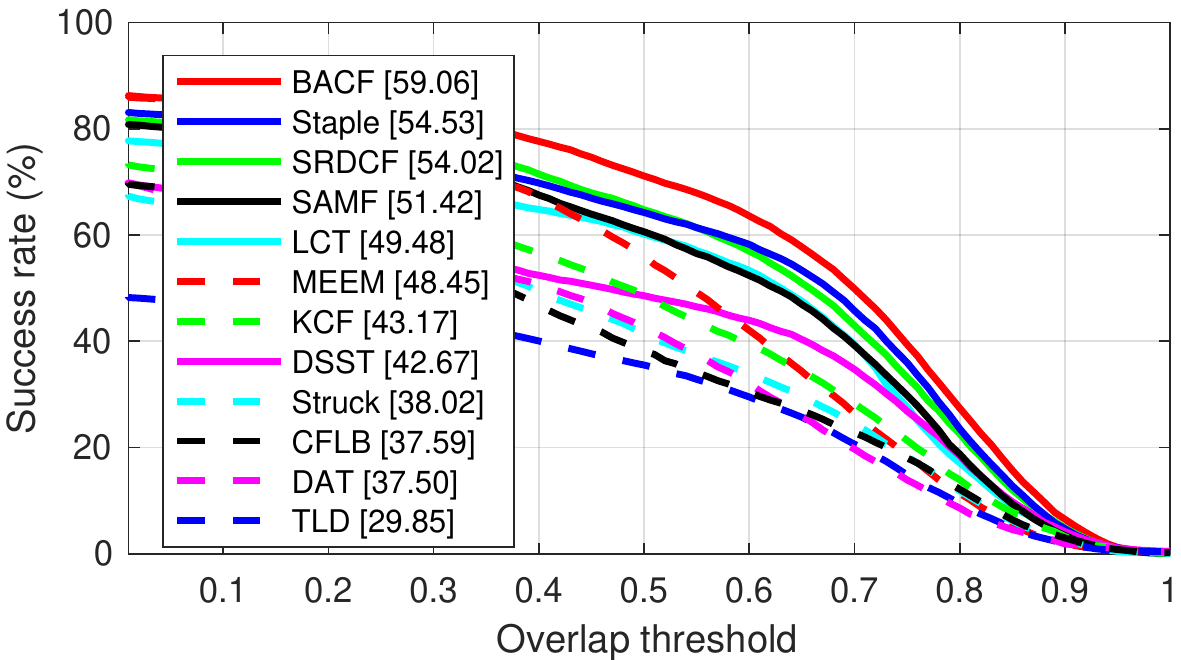} 
&  \includegraphics[scale=.45]{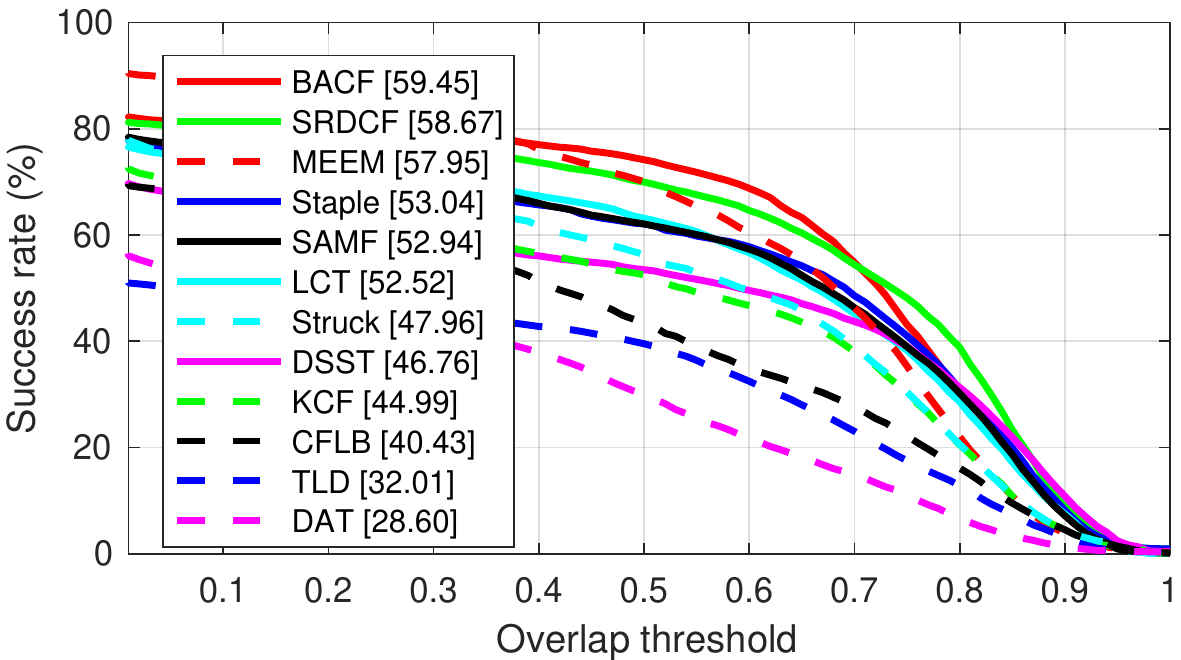} 
& \includegraphics[scale=.45]{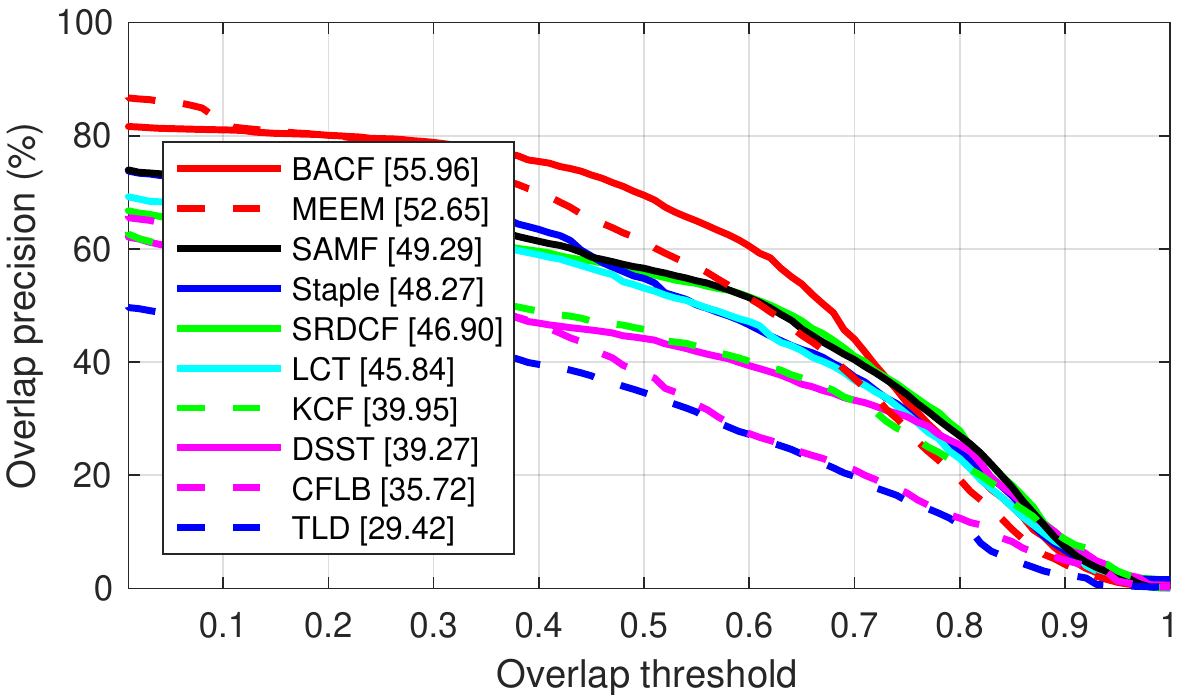}


         

             \end{tabular}
     \end{center}
            \caption{Attribute based evaluation. Success plots compare BACF with state-of-the-art HOG based trackers on OTB100. BACF outperformed all the trackers over all attributes. AUCs are reported in brackets. The number of videos for each attribute is shown in parenthesis. }
\label{Fig:Attr_OTB_100}
\end{figure*}

\begin{figure}
    \begin{center}
         \begin{tabular}{@{}c@{}  }
     {\footnotesize {\textbf{Success plot of SRE on OTB100}}} \\
     \includegraphics[scale=.485]{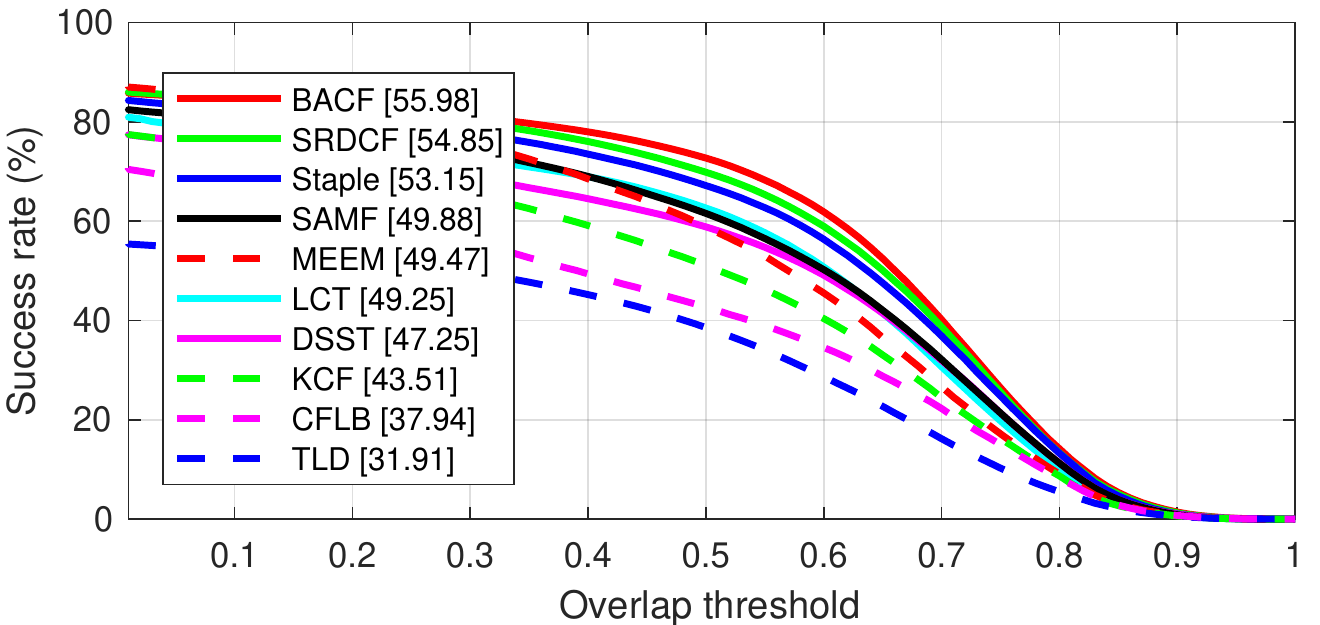} \\
    {\footnotesize {\textbf{Success plot of TRE on OTB100}}} \\
 \includegraphics[scale=.485]{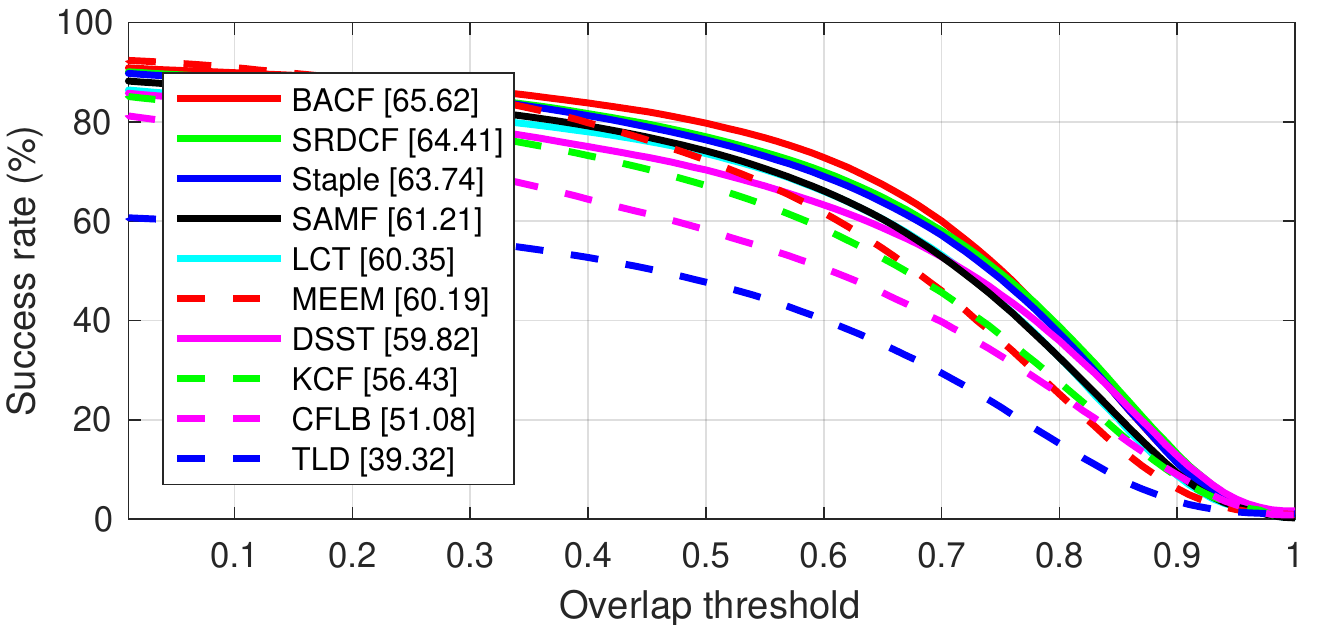} 
             \end{tabular}
     \end{center}
            \caption{SRE TRE evaluation: Success plots comparing BACF with state-of-the-art HOG based trackers on OTB100.}
\label{Fig:SRE_TRE_OTB_100}
\end{figure}

\subsection{Comparison with HOG-based Trackers}
Fig.~\ref{Fig:OP_evaluation} and Table~\ref{Table:CF_HOG} compare the BACF method with the state-of-the-art HOG-based trackers on the OTB50, OTB100 and TC128 datasets, where our method achieved the highest accuracy over all three datasets. More particularly, BACF achieved the best AUC (67.78) on OTB50 followed by LCT (62.48) and SRDCF (62.35). On OTB100, BACF (62.98) outperformed SRDCF (60.13) and Staple (58.03). BACF (51.97) is also the winner of the comparison on TC128, which is closely followed by SRDCF (51.66) and Staple (51.01). This result demonstrates the importance of utilizing background patches to learn more robust CF trackers from hand-crafted features. This evaluation also shows that BACF's strategy is more efficient than that of SRDCF to learn robust CF trackers from background. This is mainly because- unlike BACF- SRDCF has two crucial parameters (the minimum value of each weight and the impact of regularizer) to compute a regularization weight for each pixel. Since these two parameters are fixed for all videos, frames and pixels, there is no guarantee of delivering optimal results for all scenarios. Table~\ref{Table:CF_HOG} reports the average success rates of all trackers (IoU $>$ 0.50) as well as their tracking speed (FPS) on CPUs. The best tracking speed belongs to KCF (173.4 FPS) followed by CFLB (87.1 FPS), DAT (60.3 FPS) and Staple (48.3 FPS). Higher speed of such trackers, however, came at the cost of much lower accuracy compared to BACF. Our method obtained the real-time speed of 35.3 FPS which is almost 10 times faster than SRDCF (the second best tracker).

\qsection{Attribute Based Evaluation} Fig.~\ref{Fig:Attr_OTB_100} illustrates the attribute based evaluation of all HOG-based trackers on OTB100. All sequences in OTB100 are manually annotated by 11 different visual attributes such as occlusion, deformation and motion
blur. We only reported the results of 6 attributes and full evaluation can be found in the supplemental material. The results show our superior tracking performance on all attributes. This empirically demonstrates how learning CFs from a huge set of background patches improves the stability of such trackers against challenging photometric and geometric variations. Trackers such as SRDCF, Staple and MEEM showed to be less robust to out of view, occlusion, deformation, respectively.

\qsection{Robustness to Initialization} Following~\cite{danelljan2015learning, danelljan2015convolutional, bertinetto2015staple}, we evaluated our tracker towards different spatial and temporal initializations~\cite{wu2015object} using two robustness metrics: spatial robustness (SRE) and temporal robustness (TRE). SRE measures the sensitivity of a tracker against noisy initialization (small perturbations from the ground truth). TRE measures the sensitivity of a tracker when initialized at different frames of the sequence. Fig.~\ref{Fig:SRE_TRE_OTB_100} shows the TRE and SRE success plots of BACF compared with the other HOG based trackers (SRE and TRE of DAT and Struck are not available). Our method achieved the best TRE and SRE AUCs followed by SRDCF and Staple. This evaluation shows that compared to other HOG based methods our method is more robust to different spatial and temporal initializations.

\subsection{Comparing with Deep Feature-based Trackers}
\begin{figure}
    \begin{center}
         \begin{tabular}{@{}c@{}}
      \includegraphics[scale=.5]{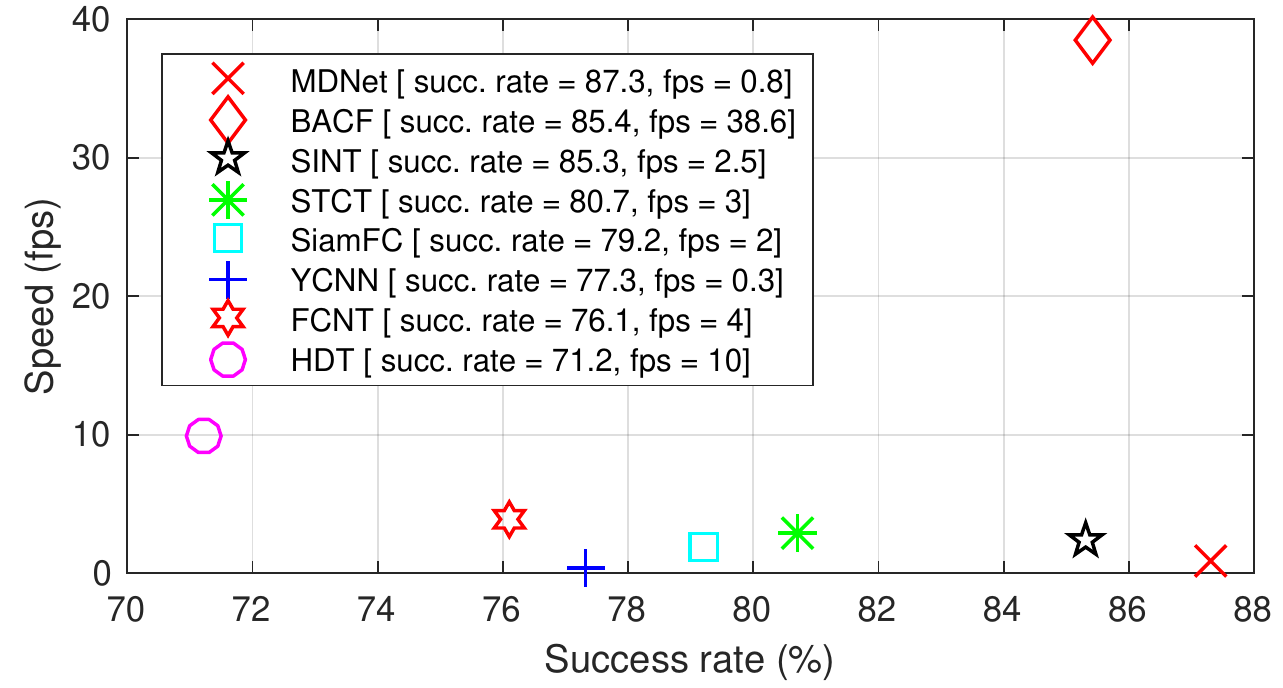} 
             \end{tabular}
     \end{center}
            \caption{Comparing our method with deep trackers in terms of tracking speed (fps) and success rate at IoU $>$ 0.50. }
\label{Fig:deep_trackers}
\end{figure}

Table~\ref{Fig:deep_trackers} compares BACF with the state of the art CF trackers with deep features, showing that BACF achieved the best success rate on OTB50, and the second best accuracy on OTB100 and TC128 after CCOT. Overall, BACF (76.0) outperformed HCF (65.0) and DeepSRDCDF (72.9) and obtained very comparable average success rate to CCOT (77.8). In terms of tracking speed, however, BACF dramatically outperformed the other trackers, with almost 170 times faster tracking speed. Despite the efficient optimization of CFs in the Fourier domain, CF trackers with deep features suffer from intractable complexity which is mainly caused by 1) computationally expensive deep feature extraction on CPUs, and 2) computing the FFT/IFFT of hundreds of deep feature channels at each frame.

\begin{figure*}
    \begin{center}
    \includegraphics[width=.9\linewidth]{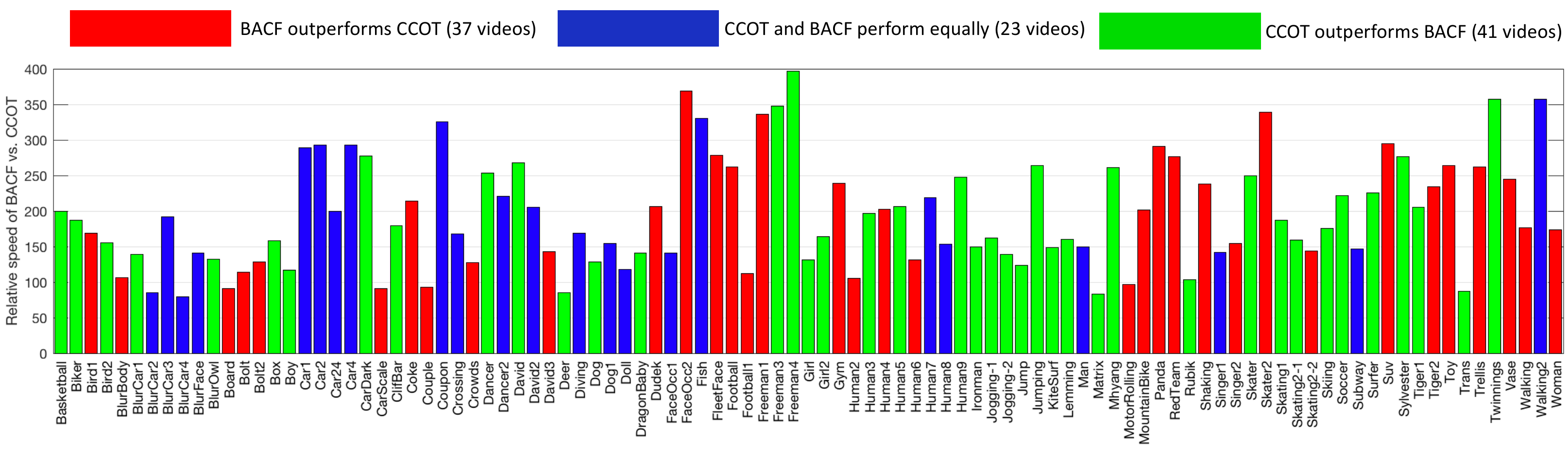}
    
    \end{center}
    \caption{Comparing BACF with CCOT in terms of speed up on OTB100. We also show performance details over which tracker performs best on a given sequence. Sequences coded red are where BACF outperforms CCOT. Sequences coded green are when CCOT outperforms BACF, and blue indicates equal performance. Across the board, we note that BACF provides an appreciable speedup (on average around 200x) compared to CCOT. CCOT, however, only outperforms BACF in terms of tracking accuracy on a small portion of the videos. For most videos, BACF performs as well or better than CCOT. }
    \label{fig:ccot_vs_bacf_1}
\end{figure*}

To emphasise the superior tracking speed of BACF and its competitive accuracy to CCOT, we directly compared these two methods on OTB100 sequences in terms of the number of videos these trackers show superior accuracy, and relative tracking speed for each sequence. Results in Fig.~\ref{fig:ccot_vs_bacf_1} show that for 37 videos (of 101 videos) BACF outperformed CCOT, and for 41 videos CCOT achieved superior accuracy. Both CCOT and BACF showed the same accuracy on 23 sequences. This comparison highlights the competitive accuracy of these methods. In terms of relative speed, BACF showed at least 100 times faster speed on all videos. For some sequences with smaller targets, such as Freeman4, FaceOcc2 and Twinnings, BACF is almost 400 times faster than CCOT.

\begin{table}
\centering
\caption{Success rates ($\%$ at IoU $>$ 0.50) of BACF compared to CF trackers with deep features. The {\color{red}first}, {\color{green}second} and {\color{blue}third} highest rates are highlighted in color. }
\label{table:CF_deep}
\begin{tabular}{lcccc}
\hline
         & BACF & HCF  & DeepSRDCF & CCOT \\ \hline
OTB50    & {\color{red}85.4} & 72.1 & {\color{blue}77.4}      & {\color{green}81.6} \\
OTB100   & {\color{green}77.6} & 64.8 & {\color{blue}76.4}      & {\color{red}81.5} \\
TC128       & {\color{green}65.2} & 58.1 & {\color{blue}64.9}      & {\color{red}70.5} \\\hline \hline
Avg. succ. rate  & {\color{green}76.0} & 65.0 & {\color{blue}72.9}      & {\color{red}77.8} \\
Avg. FPS & {\color{red}35.3} & {\color{green}0.8}  & {\color{blue}0.4}       & 0.2 \\ \hline
\end{tabular}
\end{table}

\begin{table*}
\centering
\caption{Evaluation on VOT2015 by the means of robustness and accuracy.}
\label{VOT_15}
\begin{tabular}{cccccccccccc}
\hline
  & Ours & S3Tracker & Struck & SC-EBT & LDP & DSST & DAT & Staple & SRDCF & DeepSRDCF & CCOT   \\ \hline
 Acc. & \textcolor{red}{0.59} & 0.52 & 0.47 & \textcolor{blue}{0.55}  & 0.51 & 0.49 & 0.44 & 0.53 & \textcolor{green}{0.56} & \textcolor{blue}{0.56} & 0.54  \\
  Rob. & 1.56 &  1.77 & 1.26 & 1.86 & 1.84 & 2.53 & 2.06 & 1.35 & \textcolor{green}{1.24} & \textcolor{blue}{1.05} & \textcolor{red}{0.82}   
\end{tabular}
\end{table*}

\subsection{Evaluation on VOT2015}

Table~\ref{VOT_15} shows the comparison of our method with the top 5 participants in the VOT2016 challenge\footnote{\url{http://www.votchallenge.net/vot2016/}}, CCOT, Staple and DSST on 60 challenging videos of VOT15. Our method achieved the best accuracy (0.56) by improving 5$\%$ of the accuracy obtained by SRDCF and DeepSRDCF. The highest robustness (0.82) belongs to CCOT, followed by DeepSRDCF (1.05) and SRDCF (1.24). Our tracker significantly improved the accuracy and robustness of the top participants of VOT2016 (S3Tracker, SC-EBT, LDP and DSST).

\subsection{Comparing with Deep Trackers} We also compared our tracker against recent deep trackers, including SiamFC~\cite{bertinetto2016fully}, MDNet~\cite{nam2015learning}, STCT~\cite{wang2016stct}, YCNN~\cite{chen2016once}, SINT~\cite{tao2016siamese} and FCNT~\cite{wang2015visual}. Results in Fig.~\ref{Fig:deep_trackers}, show our superior real-time performance. Surprisingly, our accuracy (85.4) is very competitive with MDNet (87.3), and our method outperformed SINT (85.3), SiamFC (79.2), STCT (80.7), YCNN (77.3) and FCNT (76.1). 
    
    \begin{figure}
    \begin{center}

\includegraphics[scale=0.67]{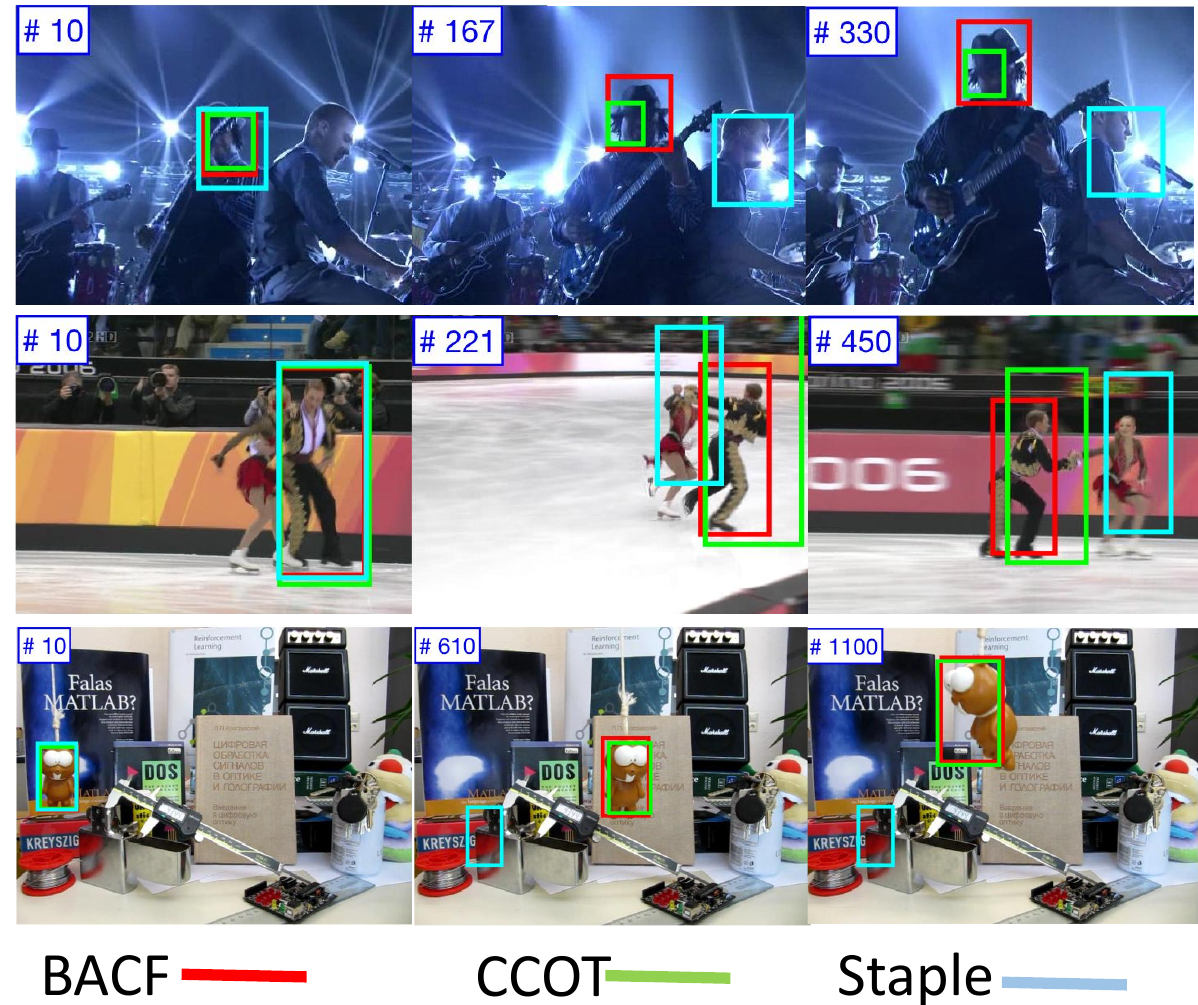}
\end{center}

          \caption{Qualitative comparison of our approach with state-of-the-art trackers on the Shaking, Skating and Lemming videos. Our approach provides consistent results in challenging scenarios, such as illumination change, fast motion and background clutter.}
\label{Fig:track_vis_good}
\end{figure}

\qsection{Qualitative Results} Fig.~\ref{Fig:track_vis_good} shows some qualitative results.

\section{Conclusion} 
In this work, we proposed background aware correlation filters for the task of visual tracking. Compared to current CF trackers which are trained by shifted patches, our method exploits real background patches together with the target patch to learn the tracker. Moreover, we utilized an online adaptation strategy to update the tracker model respect to the new appearance of the target and background over time. Learning from real patches over online adaptation significantly improved the robustness of our method against challenging deformation, scaling, and background clutter. We demonstrated the competitive accuracy and superior tracking speed of our method compared to recent CF-based and deep trackers over an extensive evaluation.

{\small
\bibliographystyle{ieee}
\bibliography{egbib}
}

\end{document}